\documentclass[sigconf,review=false, nonacm]{acmart}

\usepackage[yyyymmdd]{datetime}
\usepackage{multirow}

\usepackage{tabularx}
\usepackage{graphicx}
\usepackage{tablefootnote}

\usepackage{dcolumn} 
\newcolumntype{d}[1]{D{.}{.}{#1}}

\usepackage{subcaption}

\usepackage{todonotes}
\let\xtodo\todo
\renewcommand{\todo}[1]{\xtodo[inline,color=green!50]{#1}}

\usepackage{graphicx}
\graphicspath{ {./figures/} }

\AtBeginDocument{%
  }

\setcopyright{acmcopyright}
\copyrightyear{2018}
\acmYear{2018}
\acmDOI{XXXXXXX.XXXXXXX}
\begin{document}

\title[Elicit Human-Understandable Robot Expressions]{An Approach to Elicit Human-Understandable Robot Expressions to Support Human-Robot Interaction}

\settopmatter{authorsperrow=3}

\author{Jan Leusmann}
\orcid{0000-0001-9700-5868}
\affiliation{
  \institution{LMU Munich}
  \city{Munich}
  \postcode{80337}
out  \country{Germany}
}
\email{jan.leusmann@ifi.lmu.de}

\author{Steeven Villa}
\orcid{0000-0002-4881-1350}
\affiliation{%
  \institution{LMU Munich}
  \city{Munich}
  \postcode{80337}
  \country{Germany}}
\email{steeven.villa@ifi.lmu.de}

\author{Thomas Liang}
\orcid{0009-0007-5064-5965}
\affiliation{
   \institution{LMU Munich}
  \city{Munich}
  \postcode{80337}
  \country{Germany}
}
\affiliation{
  \institution{University of Illinois Urbana-Champaign}
  \state{Illinois}
  \country{USA}
}
\email{ttliang2@illinois.edu}

\author{Chao Wang}
\orcid{0000-0003-1913-2524}
\affiliation{
  \institution{Honda Research Institute EU}
  \city{Offenbach am Main}
  \postcode{}
  \country{Germany}}
\email{}

\author{Albrecht Schmidt}
\orcid{0000-0003-3890-1990}
\affiliation{%
  \institution{LMU Munich}
  \city{Munich}
  \postcode{80337}
  \country{Germany}}
\email{albrecht.schmidt@ifi.lmu.de}

\author{Sven Mayer}
\orcid{0000-0001-5462-8782}
\affiliation{%
  \institution{LMU Munich}
  \city{Munich}
  \postcode{80337}
  \country{Germany}}
\email{info@sven-mayer.com}

\renewcommand{\shortauthors}{Leusmann et al.}

\begin{abstract}
Understanding the intentions of robots is essential for natural and seamless human-robot collaboration. Ensuring that robots have means for non-verbal communication is a basis for intuitive and implicit interaction. For this, we contribute an approach to elicit and design human-understandable robot expressions. We outline the approach in the context of non-humanoid robots. We paired human mimicking and enactment with research from gesture elicitation in two phases: first, to elicit expressions, and second, to ensure they are understandable. We present an example application through two studies (N=16 \& N=260) of our approach to elicit expressions for a simple 6-DoF robotic arm. We show that it enabled us to design robot expressions that signal curiosity and interest in getting attention. Our main contribution is an approach to generate and validate understandable expressions for robots, enabling more natural human-robot interaction.
\end{abstract}

\ccsdesc[500]{Human-centered computing~Human computer interaction (HCI)}

\keywords{human robot interaction, gesture elicitation, robot expressions}

\begin{teaserfigure}
\centering
    \includegraphics[width=\textwidth]{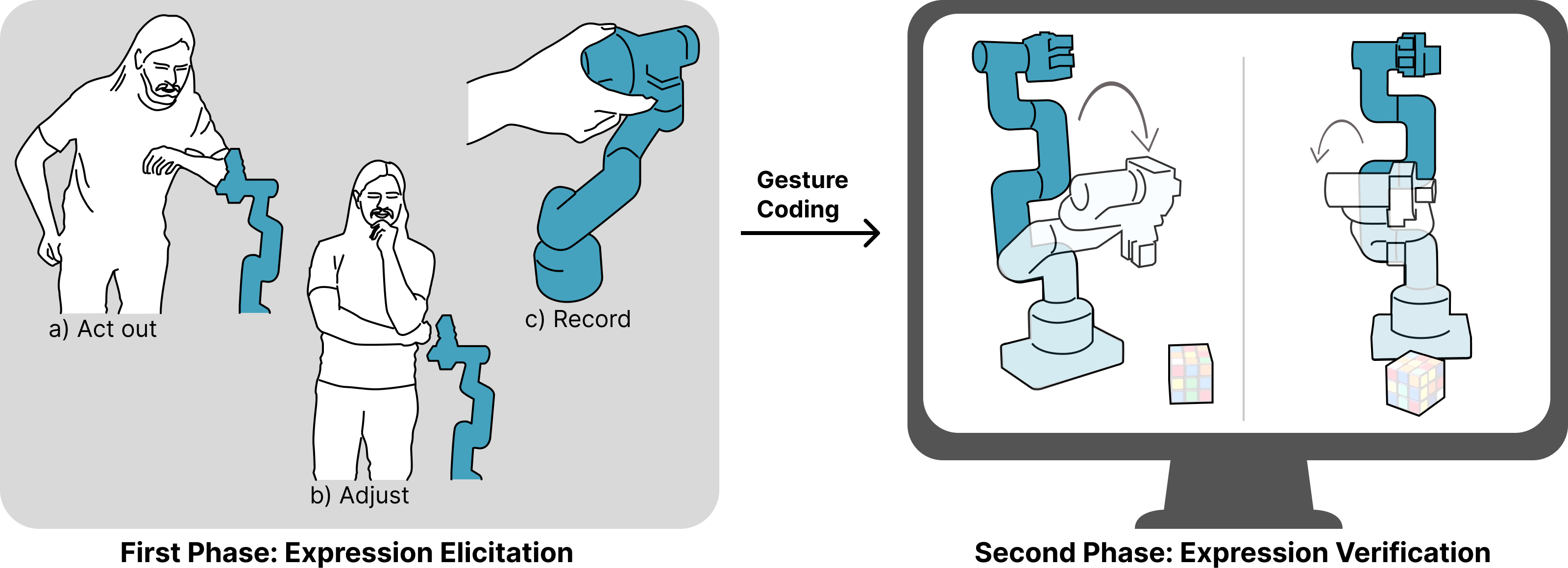}
    \caption{Two-phase process for eliciting and verifying gestures}
    \Description{The image depicts the two-phase of our new approach to create understandable robot expressions. It shows the two phases: Expression Elicitation and Expression Verification. In the first phase, called “Expression Elicitation,” a person is shown interacting with the robotic arm by acting out a gesture. The human adjusts both their posture and the robotic arm to refine the gesture. An arrow labeled “Gesture Coding” leads to the second phase, called “Expression Verification.” In this phase, the robotic arm is shown on a computer screen performing the gestures.}
    \label{fig:teaser}
\end{teaserfigure}

\maketitle

\section{Introduction}
Robots may become everyday companions, assisting with all types of tasks, including those currently beyond our capabilities. This propels us into a new era of human-robot coexistence, where robots implicitly become integral parts of our lives. Central to this is establishing effective communication between humans and robots~\cite{unhelkar2017challenges, kiesler2005fostering, sheridan1997eight}. While verbal communication is a cornerstone of human-human interaction~\cite{fer2018verbal}, humans use non-verbal cues for conveying emotions, intents, and information~\cite{reddy2012non, guerrero2016nonverbal, nikolaidis2018planning}. Non-verbal cues can either replace verbal responses or support verbal communications~\cite{phutela2015importance}. There are many existing approaches to creating robot expressions, yet, they focus on the expression creation process rather than the understandability of the expressions~\cite{jung2017affective}. However, in human-robot interaction (HRI), the expressions performed by the robot must be understandable by users to support interaction. Yet, there is no approach that focuses on creating human-understandable expressions to support human-robot interaction. 

Previous research has shown that matching verbal and non-verbal communication from robots can improve users' perception of the robot~\cite{salem2013err}. Non-verbal communication aids in conveying intentions, preferences, and emotional states, facilitating more meaningful and intuitive interactions~\cite{saunderson2019how} and positively influences both efficiency and robustness in collaborative tasks~\cite{breazeal2005effects, yang2021improving}. Many of them are in the domain of emotions~\cite{embgen2012robot, sidner2005explorations, saunderson2019how, kendon2004gesture}. 
\citet{jung2017affective} argue that when designing expressions for robots, there needs to be a shift from making the robot ``emotional'' to an understanding of how people interpret that behavior. With current methods, this is rarely the case. While we see that these expressions are created with a wide range of different methods (experts in specific fields~\cite{anderson-bashan2018greeting, press2023humorous}, motion tracking~\cite{hoffman2014designing, yamane2010animating}, or animation theory~\cite{hoffman2014designing}), applying these methods in rapid human-centered design cycles is restrictive and non-practical and ensuring human understandability not directly possible. In contrast, user-defined gestures~\cite{wobbrock2009Userdefined} can be generated rapidly by ordinary users and are widely used in human-computer interaction (HCI)~\cite{villarreal-narvaez2020systematic, vogiatzidakis2018gesture, silpasuwanchai2015designing, ruiz2011userdefined}. Gesture elicitation studies are effective in creating usable gestures by having participants create novel gestures based on given referents.  However, transferring user-defined gestures directly to robots is rarely possible due to the different capabilities. 

We propose a new approach with two phases (\textit{Expression Elicitation} and \textit{Expression Verification}) to create validated human understandable robot expressions based on prior approaches. We envision that in the \textit{Expression Elicitation} phase, participants have first to use their own bodies to imagine how they would express something given \textit{referents} through non-verbal cues and second to relay their human cues to a specific robot by performing, recording, and re-watching the newly created expressions (given the robot's form and size variations). Inspired by \citet{wobbrock2009Userdefined}, we then propose to cluster the resulting expressions, describe them using a taxonomy, and calculate occurrence scores for the resulting \textit{expressions} to generate potential \textit{expressions} for the selected robot configuration. In the \textit{Expression Verification} phase, we then use a confirmatory approach to verify the human understandability of the expressions through participants' interpretation of the displayed robot expression. We showcase our approach by creating human-understandable expressions allowing a 6-DoF robotic arm to appear curious. We conducted an example study using our approach. In the first phase, we elicited 128 robotic expressions based on eight \textit{referents}. Using open and axial coding, we coded 13 distinct expressions. In the second phase, we ran an online survey (N=260) to verify that the 13 expressions are human-understandable using videos. 

In this work, we present our approach to eliciting and validating human-understandable robot expressions. We outline the approach and showcase it with an example showcase and found that having users create robot expressions leads to human-understandable expressions even without humanoid traits. Thus, we conclude that our approach is suited to design non-verbal expressions for robots. With this paper, we contribute an approach that can improve non-verbal human-robot communication. With this, researchers and practitioners can create expressions for various robots and topics, thus understanding commonalities and differences in understandable non-verbal communication. 

\section{Related Work}
Here, we assess existing literature on non-verbal communication in human-human and human-robot interactions, as well as gesture-elicitation studies. 

\subsection{Non-verbal communication in Human-Human Interaction}
Non-verbal communication is a vital component of interaction, as 66\% to 87\% of information in face-to-face interaction is conveyed this way and often overshadows verbal communication~\cite{guerrero2016nonverbal}. We use it to communicate emotions, convey interpersonal attitudes, and manage conversations~\cite{lafrance1978cultural}. It also plays a significant role in maintaining long and effective interpersonal relationships~\cite {tiwari2016nonverbal}. Non-verbal communication includes all bodily and facial expressions~\cite{gelder2014perception}. Gestures are one vital part of non-verbal communication~\cite{goldinmeadow2013gesture}. Non-verbal communication is an essential part of human-human interaction, and especially gestures have a huge potential to express intentions and to show attention while listening~\cite{goldinmeadow1999role} and can often convey even unspoken thoughts~\cite{goldinmeadow2013gesture}. Additionally, humans are also good at understanding the gestures from other people subconsciously~\cite{gelder2014perception, dael2013perceived, wermelinger2020understand}.

However, interpreting the meaning of others' gestures is subjective and multifaceted~\cite{mcneill2005gesture}. Often, specific gestures only make sense given a specific context and can be interpreted differently in a different context~\cite{lascarides2009formal}. For example, nodding can be used to confirm, agree, submit, or give permission~\cite{poggi2010types} and, thus, also be interpreted to have these different meanings given a different context. Consequently, when asked what a gesture means without giving context multiple answers can be correct.

\textit{Non-verbal expressions enhance conversational flow, offering a simple, non-intrusive, and easily understandable means of communication. People interpret gestures within the context of ongoing conversations and recognize their nuanced meanings. Given the significance of non-verbal communication in human-human interactions, we advocate for a similar level of importance in human-robot communication.}

\subsection{Human-Robot Communication}
In the context of HRI, it also has been shown that verbal and non-verbal communication is possible in both directions~\cite{mavridis2015Review}. Similar to humans, robots can use expressions to convey their emotional state~\cite{embgen2012robot}. For example, \citet{deshmukh2018shaping} found that changing the speed and amplitudes of these gestures affect user perception in terms of Animacy, Anthropomorphism, Likeability, and Perceived Safety, measured through the Godspeed scale~\cite{bartneck2009measurement}. 
\citet{cabibihan2012HumanRecognizable} conducted a study with a humanoid robot mimicking humans and found that humans, in general, understand what the robot tried to express. Robots using gestures increase the willingness of users to engage with them~\cite{si2016using}. It has been found that matching gestures and verbal output increased participants' likeability of the robot and willingness to interact~\cite{salem2013err, kim2012robot}. Humans are more willing to approach robots that use gestures~\cite{si2016using}. In collaborative tasks, robot gestures increase communication and understanding between the two parties and can even lead to the users perceiving a lower workload~\cite{lohse2014robot}.

Researchers have investigated how to improve interaction with non-humanoid robots. Although humanoid robots are often the way how we envision future robots to look like due to their higher levels of anthropomorphism, non-humanoid robots will be an essential part of future interaction. Due to the various areas of application for robots, it is often necessary to choose a robotic design fitting into the environment, e.g., a humanoid robot supporting us with kitchen tasks would stand in our way, but a robotic arm from the ceiling could achieve the same results without being in the way~\cite{oechsnerchallenges}. Even slight changes in the appearance of robotic arms can lead to higher levels of anthropomorphism~\cite{terzioglu2020designing}. Furthermore, humans perceive even inanimate objects as social entities, attributing emotions, inner states, and personality to them~\cite{koppensteiner2011perceiving, reeves1996media}. Non-humanoid robots are able to express emotions~\cite{hoggenmueller2020emotional}, the will to collaborate~\cite{cha2016nonverbal}, and can be used to support interaction between humans~\cite{press2023humorous, hoffman2015design} using gestures. It is crucial for robots in the various aspects of our lives to communicate their intent so that humans can make more informed decisions about their interaction with them~\cite{ziemke2020understanding}. While these studies offer solutions for expressive robotic gestures, not much is directed toward curiosity, and many of the gestures are developed based on existing research on human gestures or even ambiguously~\cite{cha2016nonverbal, embgen2012robot, kim2012robot, laplaza2022ivo, press2023humorous}, which is generally harder to implement on a non-humanoid robot. In our example study, we focus specifically on the expression of curiosity by a robot and generate our gestures through an elicitation phase.

\textit{Effective communication is essential in HRI, particularly in collaborative scenarios where mutual understanding is crucial. Robots that employ gestures positively impact collaborative tasks. Therefore, enhancing HRI involves using non-verbal communication that humans can implicitly understand.}

\subsection{Expressions in Human-Robot Interaction}
\label{sec:gestureshri}
Expressions are a means of communication between humans and robots, as they play a vital role in facilitating effective and intuitive interaction between two parties~\cite{sidner2005explorations, saunderson2019how}. They can convey information, intentions, and emotional states to humans. By mimicking and utilizing human-like gestures, robots can enhance their ability to communicate and engage with humans in a more natural and intuitive manner. Non-verbal communication aids in conveying intentions, preferences, and emotional states, facilitating more meaningful and intuitive human-robot interactions~\cite{saunderson2019how}. They can make robots appear more polite~\cite{kumar2022politeness}, improve task performance in human-robot collaboration~\cite{breazeal2005effects}, and convey affection through changes in proxemics, color, sound, and kinematic properties, which affects the emotional valence of users~\cite{bethel2008survey, tan2016happy}. Most current studies focus on designing movements that convey emotions and whether humans can correctly interpret these emotions rather than examining the gestures classified by \citet{kendon2004gesture}.

The creation of expressions for robots is currently performed with various different methods~\cite{wolfert2022review}. Currently, expressions are being created through automatic pipelines~\cite{deng2018Generative, kim2007Automatic, suguitan2023face2gesture}, through combination and mapping of predefined motion sets~\cite{vandeperre2015Development, yang2014Robotic}, the usage of motion capturing data~\cite{rosado2014KinectBased, rosado2014Using, boutin2010Autoadaptable, heloir2006Captured, nakazawa2002Imitating}, in applicable fields (e.g., humor, animation, choreography) expert knowledge and design processes~\cite{press2023humorous, anderson-bashan2018greeting}, a combination out of experts and motion capturing data~\cite{reiley2010Motion}, or animation design theory~\cite{hoffman2015design, ribeiro2020Practice}. Especially methods using motion-capturing data often imitate actual human gestures, proposing that robots should communicate non-verbally exactly like humans. Generating this data or gaining knowledge from experts is often costly or unfeasible. 

\textit{In summary, expression creation typically relies on motion tracking data, expert knowledge, or animation theory, which can be time-consuming and costly. Furthermore, these methods may not guarantee that humans easily understand the resulting expressions. In general, qualitative studies are underutilized in HRI research~\cite{hoffman2021primer}.}

\subsection{Gesture Elicitation}
With the seminal work by \citet{wobbrock2009Userdefined}, user-defined gestures for human-computer interaction became the norm for designing gesture inputs for today's computing systems. Recent literature reviews found that over 267 gesture elicitation studies have been conducted~\cite{villarreal-narvaez2020systematic, villarreal-narvaez2024brave}. Although originally planned for eliciting hand gestures as input for interactive systems, the core idea of gesture elicitation studies can be used with different body parts and in other application areas~\cite{fan2017empirical, chamunorwa2023empirical}. In gesture elicitation studies, many participants create multiple single gestures, given a referent. Originally, a referent is one feature for which the participants should create a new gesture. The method then proposes to merge similar gestures, calculate an agreement score for the gestures for each referent, and then choose one different gesture with the highest agreement score for each referent. On average, studies use 16 referents and collect 723 gestures from 25 participants~\cite{villarreal-narvaez2024brave}. 

\textit{So far, gesture elicitation studies have mainly been used to map gestures to device functionalities and not to create robot gestures. This approach can potentially eliminate the need for experts, motion data, or specialized animation theory. While gesture elicitation studies are effective for generating gestures in new interaction concepts, they are not entirely suitable for our purposes as robot gestures do not require distinct variations for different statements. However, it can spark new directions for expression elicitation.}

\section{Research Approach to Elicit Robot Expressions: A Research Plan}
\begin{figure}[t]
   \centering
   \includegraphics[width=1\linewidth]{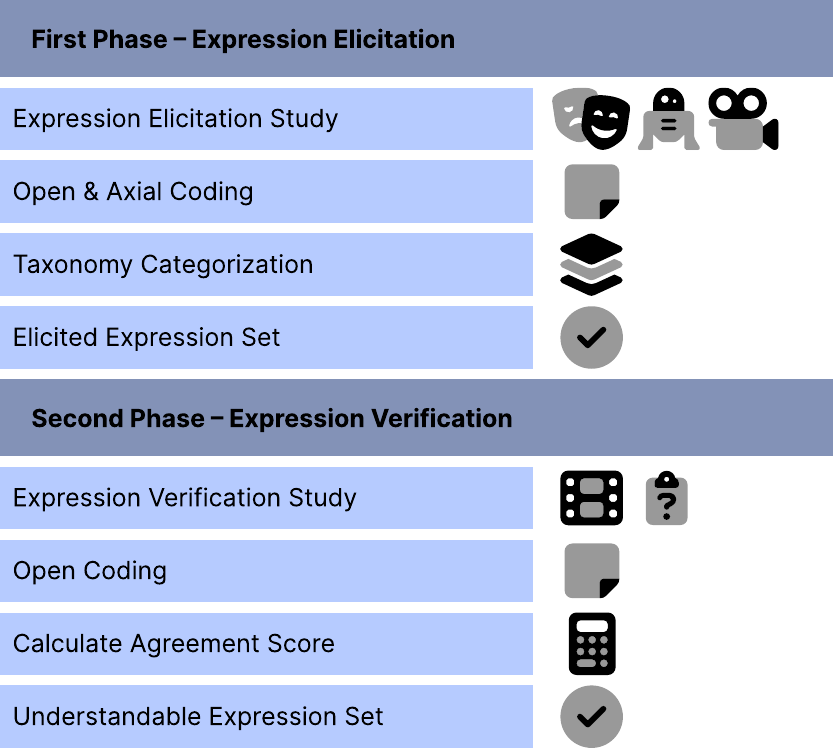} 
   \caption{Procedure of creating human-understandable robot expressions, divided into two phases; \textit{Expression Elicitation} and \textit{Expression Verification} leading to a final understandable expressions set.}
   \Description{The figure describes the procedure of our proposed research approach. It reads from top to bottom, and depicts the 4 stages for each of the two phases. Phase 1 - Expression Elicitation has the four stages: Expression Elicitation Study, Open and Axial Coding, Taxonomy Categorization, Final Expression Set. Phase 2 - Expression Verification has the four stages: Expression Confirmation Study, Open Coding, Calculate Agreement Score, and Understandable Expression Set.}
   \label{fig:procedure}
\end{figure}

We envision a new research plan to create human-understandable expressions for robots easily and in a user-centered way, see \autoref{fig:procedure}. This method uses a two-phase approach and can be applied to multiple domains where creating understandable robotic expressions is required. 

\subsection{Approach Overview}
Human natural expressions are generally human-understandable~\cite{kendon2004gesture}. However, not all robot form factors can directly mimic human gestures as they do not resemble typical human bodies. To bridge this gap, we employ a two-phase approach containing an \textit{Expresion Elicitation} and \textit{Validation} phases. First, we conduct an \textit{Expression Elicitation} phase, where participants enact expressions with their own bodies, explaining their thought processes. Then, they create corresponding robot expressions, keeping their physical gestures in mind. We propose to ensure that robot movement manipulation is user-friendly, allowing participants to adjust joint configurations manually and easily. Afterward, participants can view and iterate over their newly created robot expression until they match their envisioned expression. 

For the expression \textit{Expression Verification} phase, we suggest to ask for the participants' unbiased opinions on what they think the robot expressed. Thus, we suggest asking participants to interpret the expression as the first thing of the phase, without the possibility of them seeing other information from the subsequent questions. In order to mitigate potential biases in the measurements, it would be recommended to choose a between-subject study design~\cite{cockburn2019anchoring, greenwald1976subjects}. 

\autoref{fig:procedure} shows an overview of the proposed approach. In detail, in our two-phase approach to creating human-understandable gestures, the first phase is strongly influenced by the well-established method of gesture elicitation from \citet{wobbrock2009Userdefined}. However, in contrast to most gesture-elicitation studies, we do not need to create a different expression for every \textit{referent}; humans can map (and do map) singular expressions to different meanings, but differentiate them with contextual information. The second phase is to confirm that an independent sample of participants correctly interprets expressions created in the first phase. Finally, we can calculate which expressions were generally perceived to express the same things as the original \textit{referent} asked about.

\subsection{Goals and Requirements}
\label{sec:approachGoals}
As robots become companions for different tasks in daily life, it is important to understand how users understand and perceive the interaction with robots. Yet, a general understanding of human-robot interaction is impacted by the wide range of robot configurations and form factors. This heavily impacts communication types that are typically linked to a specific body configuration, such as human expressions of curiosity, for example, as we as humans can likely recognize. However, it is unclear how this bodily communication would extend to the different robot form factors available. To address these challenges, we derived a set of goals and requirements for developing human-understandable robot expressions that at the same time, is generalizable:
\begin{enumerate}
    \item \textbf{Universal}: The method should be universally applicable, working seamlessly with different robot types.
    \item \textbf{Generalizable}: The method should enable the generation of expressions for any topic or domain.
    \item \textbf{Accessible}: Researchers, regardless of their available resources, should find the expression creation process straightforward, efficient, and capable of yielding human-understandable expressions. 
    \item \textbf{Comparable}: The results should be standardized and comparable across all studies utilizing this approach, facilitating a comprehensive analysis of the key elements in an expression's effect.
    \item \textbf{Validated}: The created robot expressions should be easily and accurately interpreted by a broad and diverse user group, ensuring their effectiveness in communication.
\end{enumerate}

\subsection{Blending Existing Approaches}
To establish a foundation for generating and evaluating expressions or gestures, we explored several existing approaches. For expression generation, we considered methods such as predefined motion sets~\cite{vandeperre2015Development, yang2014Robotic}, motion capturing data~\cite{rosado2014KinectBased, rosado2014Using, boutin2010Autoadaptable, heloir2006Captured, nakazawa2002Imitating,reiley2010Motion}, expert knowledge~\cite{press2023humorous, anderson-bashan2018greeting}, or animation design theory~\cite{hoffman2015design}. In contrast, HCI research often favors user-centered techniques like gesture elicitation studies~\cite{villarreal-narvaez2020systematic}, focus groups~\cite{karat2003evolution}, mimicking~\cite{celalettin2016imitation}, or role-play~\cite{morozlapin2009role, asensio2011use}. We draw from their insight to design the first phase of our approach: \textit{Expression Elicitation}. In traditional gesture elicitation studies, users are asked to generate gestures they want to perform on a device to get a certain output. The idea is that if many users come up with a similar idea of how a gesture for one action should look like, this is an intuitive gesture. We use this idea to let users create gestures that they want the robot to perform, instead of themselves. The intuition is the same; if many users agree that one robotic gesture means a certain thing, that gesture is an intuitive gesture.

As \citet{wobbrock2009Userdefined} highlighted in their seminal paper, validation is an important next step after elicitation. This aligns with other methods like scale development~\cite{boateng2018best}. However, it is not always common practice to validate new sets~\cite{villarreal-narvaez2020systematic}. Consequently, we build the verification set directly into our approach as the second phase: the \textit{Expression Verification}. Here, we propose to use a large set of participants and ask, without priming them, what they think a robot wants to express doing one certain expression. If the large majority of participants understand one expression with the same meaning, this expression is a reliable way to communicate the goal, with which the expression was originally created. 

\subsection{Choice of Participants}
As bodily expressions are often part of implicit memories, which the individual cannot explicitly articulate in words or descriptions, We envision that during the  \textit{Expression Elicitation}, participants interact physically with the robots, allowing them to manipulate the body of the robot, or to use their own bodies as a tool to articulate the expressions. Thus, the phase should ideally be conducted as a lab study or in situ study. Depending on the explicit context for which our method should be used to create new expressions, both domain experts and regular users could be included in the elicitation process. Domain experts have deeper knowledge and background, leading to more informed opinions. However, non-verbal expressions are inherently human; thus, regular users also have an understanding of how they would express certain concepts. 

As we propose that the first phase needs to be an in-person lab study, this limits the diversity of the participant pool. Thus, we propose to counteract the second phase as an online survey. We propose the evaluation and verification of the understandability of the expressions created in the first phase using a diverse large-scale study. In an online survey, video recordings of the expressions will be necessary to convey the newly designed expressions to the participants. 

\subsection{Choice of Referents}
In gesture elicitation studies, a \textit{referent} is the target or subject of the designed gestures~\cite{wobbrock2009Userdefined}. In these traditional studies, researchers aim to understand the relationship between a \textit{referent} and gestures to design more effective and user-friendly interfaces. In traditional gesture elicitation studies, each \textit{referent} has to have one distinct gesture associated with it. This contrasts our case; here, the referents are used to generate expressions for specific topics. However, they do not need to be unique, as in human-human communication, we often use the same non-verbal communication patterns for multiple different interactions, and the conversation partner can infer from the context what is meant by that expression. Thus, the choice of referents does not need to reflect the exact set of expressions one wants to create with this approach. Yet, the referents serve as a possibility for designers to instruct users to allow two design expressions with a use case in mind. In detail, each referent should prompt the user to perform a non-verbal expression for a certain topic. We suggest using additional control referents. The aim is to create expressions that are unrelated or contrasting to the domain of created expressions for verification.

\subsection{Analysis and Taxonomy: Finding Commonalities in Expressions}
To condense findings from the expression elicitation phase, we suggest a two-step coding process for the robot expressions created by participants in the first phase. First, we recommend using open coding to identify commonalities and subtle differences among expressions, preserving as much expression information as possible. Next, we propose clustering similar expressions by referent through axial coding, with the possibility of similar clusters emerging across different referents. Initially, this will yield numerous unique expressions, each described in detail using our taxonomies. However, in subsequent stages, we advocate for condensing these expressions into their core ideas. To achieve this, we suggest clustering groups by their primary components into what we term \textit{expressions}. Once again, the same \textit{expressions} may appear across different \textit{referents}. From this point forward, we suggest analyzing the \textit{expressions} independently of the \textit{referents}. These \textit{expressions} should clearly convey some internal state of the robot. We propose the following set of essential properties that describe each \textit{expression} in great detail, see \autoref{tab:attributes}. This makes it easier to decide which expressions are the same and, thus, need to be merged or which are different from each other. However, depending on the robot used, these taxonomies can differ, as these properties should mainly be there to provide an easier but also more transparent way of coding the expressions. To describe and distinguish the emerged \textit{expressions}, we propose to include, at least, these six taxonomies: 

\textbf{Speed} provides a subjective assessment of expression speed. The default speed is labeled as "normal," while expressions significantly faster or slower than this default are classified as ``fast'' or ``slow.''

\textbf{Complexity} describes whether an expression consists of a single motion or a combination of multiple motions. An expression is considered ``compound'' if it involves multiple sequential movements or simultaneous combinations of different movements (e.g., vertical face movement while simultaneously moving the body from left to right).

\textbf{Flow} classifies whether an expression involves continuous, uninterrupted motion or discrete, non-continuous motion. Compound expressions may combine both continuous and discrete movements.

\textbf{Binding} assesses whether the robot's face is anchored toward a specific object or person and moves independently.

\textbf{Dynamics} describe the robot's movement and should be further specified in the context of the expression's target (e.g., moving away) and the robot's configuration.

\textbf{Focus} characterizes the relationship between the communication node and the target. The communication node refers to the point a user would identify as the robot's ``face'' (e.g., the gripper for robot arms or the head for humanoid robots).

\begin{table}[b]
\centering
\caption{Taxonomy for human-understandable robot \textit{expressions}. The taxonomy includes six dimensions.}
\label{tab:attributes}
\begin{tabularx}{\linewidth}{llX}
\toprule
\textbf{Dimension} & \textbf{Category} & \textbf{Description} \\
\midrule
\multirow{3}{*}{Speed} 
& Slow & Expressions' speed is slow\\
& Normal & Expressions' speed is neither slow nor fast\\
& Fast & Expressions' speed is fast\\
\midrule
\multirow{2}{*}{Complexity}& Single & Expression is one motion\\
& Compound & Expression consists out of multiple motions\\
\midrule
\multirow{3}{*}{Flow} 
& Continuous & Expression is a continuous motion\\
& Discrete & Expression is a discrete motion\\
& Combined & Expression is a combination of discrete and continuous motions\\
\midrule
\multirow{3}{*}{Binding} 
& Environment & Expression is environment-oriented\\
& Object & Expression is object-oriented\\
& Person & Expression is person-oriented\\
\midrule
\multirow{2}{*}{Dynamics} 
& Dynamic & Robot moves significantly for the expression to a specified direction\\
& Static & Robot stays mostly static\\
\midrule
\multirow{2}{*}{Focus} 
& Focused & Expressions focus is anchored on its target\\
& Unfocused & Expression does not have an anchored target\\
\bottomrule
\end{tabularx}
\end{table}

\subsection{Measurements}
For measurement purposes, we recommend employing established domain-specific questionnaires whenever feasible in both study phases. In the first elicitation phase, these questionnaires serve to gauge participant satisfaction with the created expressions and identify anomalies. More importantly, the questionnaire should be integrated into the second expression verification phase, with results reported. Additionally, we suggest asking about the user perception of the chosen expressions using a subset of questions proposed by \citet{rzayev2019Notification}, as this can provide additional properties of each expression. Lastly, and most importantly, participants of the second phase should be asked what they think the robot expressed without biasing them with any previous domain information. However, depending on the domain for which the expressions should be created, a certain context should be given to the participants, as the same expression can mean different things in different contexts.

\subsection{Metrics}
\label{sec:metrics}
We propose one metric each for the \textit{Expression Elicitation} and \textit{Expression Verification}. In the first phase of our approach, we want to determine which gestures are suggested by how many participants. With the \textit{occurrence score}, we can later evaluate whether expressions suggested by more participants are also better understood. In the second phase, we want to measure how much participants agree on one expression expressing one specific thing. For this, we propose calculating the \textit{qualitative response accuracy}.

To measure the agreement for the \textit{Expression Elicitation}, the agreement score (A)~\cite{wobbrock2009Userdefined} and agreement rate ($\mathcal{AR}$)~\cite{vatavu2015formalizing} do not apply to our case. The reasoning for this is three-fold: (1) We do not need to find one extremely well-fitting expression for one \textit{referent} but rather find out which expressions are understood in which way. The same expression can be the best fit for multiple \textit{referents}, and we encourage diversity in expressions. (2) The results of this method should state how confident humans understand each expression. (3) Users can propose multiple candidates (this is a common issue~\cite{morris2012Web}). However, $\mathcal{AR}$ penalizes diversity due to squaring the ratio. For this reason, we favor a linear relationship to calculate a quality metric. For this, we propose to rate each expression in a given set of expressions of one \textit{referent} by an occurrence score ($OS$):
\begin{equation}
\label{eq:os}
    OS_{R_i,E_j} = \frac{|E_j|}{|R_i|}, E_j \in R_i,
\end{equation}
with which can calculate the OS for all expressions $E_j$ per \textit{referent} $R_i$. An $OS$ of 1 describes that for a given \textit{referent}, only one expression was envisioned by participants, and a score of $\frac{1}{N}$ denotes the lowest possible score, where $N$ distinct expressions have been proposed once.

An additional approach to understanding the diversity and elicitation saturation is the \textit{consensus-distinct ratio}, which was introduced by \citet{morris2012Web} for gestures. Here, the idea is to count all gestures that are named more often than a given threshold. Here \citet{morris2012Web} proposes a default threshold of $\geq 2$. We argue that this metric scoring low could have two potential reasons. (1) The saturation point of gestures has not been reached. Thus, more participants are needed to fully understand potential expressions for this \textit{referent}. (2) The \textit{referent} itself lacks specificity, and thus, consensus is hard to reach. Other metrics, as \textit{max-consensus}~\cite{morris2012Web} or $\mathcal{CR}$~\cite{vatavu2015formalizing} are not applicable to evaluate our case of expressions as they tend to find well-separable outcomes, which we do want to force onto our resulting set of expressions. 

In line with \autoref{eq:os}, we propose to quantify the perceived \textit{qualitative response accuracy} ($QRA$) for the \textit{Expression Verification}. Here, to calculate $QRA$ by textual responses from participants who were asked to describe the expression. Through open coding, the researchers then need to determine if the response matches the intended meaning of the \textit{expression} while respecting the meaning of the original \textit{referent}. Here, we count the fitting labels as $C^+$, and the not-fitting ones as $C^-$. As a result of respecting the initial \textit{referent}, an \textit{expression} can have multiple $QRA$'s. As such, we calculate $QRA$ as:
\begin{equation}
\label{eq:qra}
    QRA_{R_i,E_j} = \frac{|C^+_{E_j}|}{|C^+_{E_j}|+|C^-_{E_j}|}, E_j \in R_i,
\end{equation}
A high $QRA$ means the human-created \textit{expression} leads to a human-understandable \textit{expression}.

\section{Example Process of the Two Phases}
In the following, we show one full cycle of our approach. First, we conducted an expression elicitation phase to create expressions to signal curiosity. Secondly, we conducted an expression verification phase to confirm whether the created expressions were understandable.

\subsection{Example Expression Elicitation Study}

\subsubsection{Study Design}
We conducted a lab-based expression elicitation study with 16 participants, allowing them to create and refine the robot expressions directly. Participants were provided with eight different \textit{referents}. They first acted out the gestures with their bodies, explaining their thought process, and then translated these expressions onto the robot. This approach helped them mentally align with how they would express each referent. We used Latin-square ordering to counterbalance the \textit{referents}' presentation order.

\subsubsection{Apparatus}
For our study, we used the MyCobot 280 M5 robotic arm from Elephant Robotics\footnote{\url{https://www.elephantrobotics.com/en/mycobot-en/}}. We chose this specific robot because it is lightweight and small, making it easy for participants to manipulate its movement with their hands. Its lack of anthropomorphic traits offers a wide range of possible degrees of freedom with its movement and does not prime participants into just copying their own, e.g., head movement.

To record and play the robot's joint positions and movements, we created a simple Python application using TKinter for the GUI, the MyCobot Python library, and ROS\footnote{ROS 1 Noetic running on Ubuntu 20.04}. The application allowed the study conductor to record key points of the robot's joint positions, shown by the participants, into one motion. The created motion could then directly be played back to the user and allowed for direct refinements. The study conductor handled the creation tool while the participants gave instructions on how they wanted the robot to move. Furthermore, we video-recorded the participants' movements and explanations for later analysis. 

\subsubsection{Referents}
We selected six curiosity-related \textit{referents} and two control \textit{referents} for our study. We found six core ideas from related work on how to express curiosity. (1) Visually intake more information in unfamiliar environments~\cite{gottlieb2014attention}, (2) Adapt sensor positioning to intake sound better~\cite{altun2019underestimated, darwin1872expression}, (3) Construct knowledge about unknown things~\cite{piaget1945play, bonawitz2012children, wang2021Children}, (4) Express interest in other people to learn more about them~\cite{altun2019underestimated}, (5) Express will to listen~\cite{altun2019underestimated}, and (6) Show attention and engagement to express interest~\cite{dael2012emotion}. From this, we constructed the six curiosity \textit{referents} (R1-R6) and two control \textit{referents} (R7, R8). \autoref{tab:referents} shows all referents.
\begin{table}[t]
    \centering
    \caption{The eight referents we used to prompt users to create expressions. We showed these referents to users and then asked them to create an expression the robot would do to match the descrption.}
    \label{tab:referents}
    \begin{tabularx}{\linewidth}{cX}
    \toprule
    \textbf{Referent} & \textbf{Description} \\ 
    \midrule
    \textbf{R1}       & Imagine a non-movable object on the table in front of you. Without touching it, how would you visually observe the object to intake more information? \\ 
    \textbf{R2}       & Imagine a constant sound in the corner of the room. How would you listen to the sound to intake more information? \\ 
    \textbf{R3}       & Imagine a non-movable object on the table in front of you, but you don't understand how it works (e.g., why a computer screen isn't black). How would you reduce uncertainty about this object? \\
    \textbf{R4}       & Imagine someone is explaining something to you. How would you show understanding and acknowledge the information? \\ 
    \textbf{R5}       & Imagine someone starts talking to you. How would you use an open posture to signal that you are listening? \\ 
    \textbf{R6}       & Imagine someone is talking to you. How would you express that you are engaged and attentive? \\ 
    \textbf{R7}       & Imagine you see something scary. How would you back away in fear? \\ 
    \textbf{R8}       & Imagine someone starts talking to you. How would you use a closed posture to signal that you are rejecting what they are saying? \\
    \bottomrule
    \end{tabularx}
\end{table}

\subsubsection{Task}
\label{sec:task}
For each of the eight \textit{referents}, we instructed participants to create an expression for the robot by first acting it out themselves and then replicating it on the stationary robot. The only constraint was that they could not move away from their position, as the robot was also mounted stationary. After acting out the referent, participants physically positioned the robot's joints to certain key points and then asked the study conductor to save the position when they were satisfied, stating their preferred duration, speed, and pauses for each movement. The tool allowed participants to preview, modify, and refine their expressions, including undoing points or adjusting the speed.

\subsubsection{Procedure}
The study began by welcoming the participants, introducing the study objectives to them, a request for their consent, and a demographics questionnaire. Then the study began with two tutorial referents (the emotions happy and sad) to explain and show participants the process of recording gestures with the robot.

For the first tutorial referent, the study conductor demonstrated the process while explaining their thought process. In the second, participants performed the task independently with guidance.
After completing the tutorials, participants moved on to the eight study referents, creating and refining one expression for each until they were satisfied.

After recording the expressions, we played them back to participants in a random order, asking them to rate each one using five questions. We formulated these questions based on curiosity literature, reflecting the pillar from which we created our \textit{referents}: ``I perceived the system to be very [engaged, attentive, explorative, information seeking, curious]''. Participants could watch the robot's performance as many times as they wanted. After rating each expression, participants moved to the next without seeing the original referents. Finally, we gathered feedback, addressed any questions, and thanked them for participating.

\subsubsection{Participants}
We recruited 16 participants (male = 10, female = 6) via convenience sampling. Their education levels included ``master's degree'' (11), ``bachelor's degree'' (3), and ``some college but no degree'' (2). Their age ranged from 20 to 35 ($M=27.50$, $SD=4.26$). Our participants had three different nationalities: German (12), USA (3), and Canada (1). The mean Affinity for Technological Interaction (ATI)~\cite{wessel2020Affinity} score across all participants was 4.61 ($SD=0.68$, Cronbach's $\alpha=0.85$). On average, the study took 45 minutes, and we reimbursed participants with 10\texteuro.

\subsubsection{Analysis}
Three authors used open coding to identify and merge similar expressions, initially coding 128 expressions and condensing them into 37 distinct expressions. We then applied axial coding to further consolidate these into 13 unique expression categories, based on how the robot moved and what it conveyed. The final expressions are documented in the Supplemental Material.

\begin{figure*}[t]
    \centering
    \includegraphics[width=\linewidth]{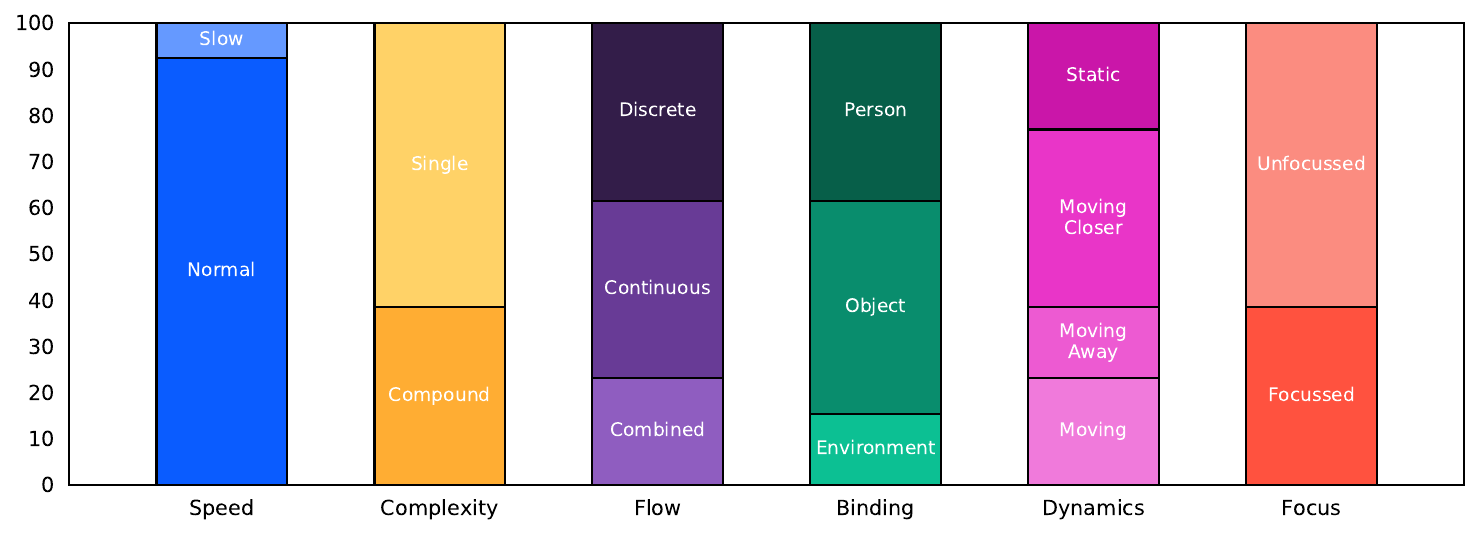}
    \caption{Distribution of expressions in six taxonomy categories (c.f., \autoref{tab:expressions}).}
    \Description{This figure displays the distribution of our expression on the six taxonomy categories. The large majority of gestures were normal speed; 60\% of gestures had single complexity and 40\% compound complexity. Flow has approximately an equal distribution between discrete, continuous, and combined. The majority of expressions had an object-binding, followed by person, and lastly environment (approximately 12\%). For dynamics, the majority is categorized as moving closer, followed by static and moving, and lastly, moving away. And lastly, approximately 60\% of gestures were unfocused and 40\% focussed.}
    \label{fig:distributiontaxonomies}
\end{figure*}
In detail, three researchers iteratively coded the robot gestures, observing that participants often perceived the robot's end-effector as a face, as reflected in their reactions and think-aloud comments. We established the front as representing the direction of the object in the referent, while other movements were categorized as body gestures (e.g., extension, contraction, leaning). After coding 12\% of expressions independently, we consolidated our approach and collaboratively coded the rest, yielding 37 expressions, which were grouped by \textit{referents}, though the same expression could appear across multiple \textit{referents}.

In the second step, we condensed 37 expressions into 13 \textit{expression categories} using axial coding. We categorized expressions based on their primary components. This process unified variations in movement, which appeared repeatedly across different referents, see \autoref{tab:expressions}. Each \textit{expression category} could be recurrent across various \textit{referents}.

\begin{table}[t]
\caption{Description of the 13 resulting robot \textit{expressions}. In referents shows which \textit{referents} created each \textit{expression}.}
\label{tab:expressions}
\small
\begin{tabularx}{\linewidth}{llX}
\toprule
Expression & In Ref. & Description                                                     \\
\midrule
E01        & R1, R3     & Getting closer with end-effector and ``scanning''\footnotemark the object in one axis.                        \\
E02        & R1, R3     & Getting closer with end-effector, continuously moving around the object and ``looking.''              \\
E03        & R1, R3     & See E02, but doing the movement multiple times sequentially ``looking'' at the object from two sides. \\
\midrule
E04        & R2         & Leaning forward, scanning environment by turning the end-effector horizontally to both sides.               \\
E05        & R2         & Learning forward, facing one spot in the environment.         \\
E06        & R2         & Repeatedly leaning forward, ``looking'' at one spot in the environment.      \\
\midrule
E07        & R4, R5, R6 & Nodding with the end-effector                                                        \\
E08        & R4, R5, R6 & Nodding with the end-effector with body/end-effector movement in between.          \\
E09        & R5, R6     & Showing attention through body movement.                           \\
\midrule
E10        & R7         & Body leaning back, ``end-effector'' focussed on a target.                           \\
E11        & R7         & Body first leaning forward then back, ``end-effector'' focussed on a target.              \\
\midrule
E12        & R8         & End-effector turning horizontally to both sides multiple times (shaking head).                                                \\
E13        & R8         & Body turning and moving away from the target.                                              \\
\bottomrule
\end{tabularx}
\end{table}
\footnotetext[3]{The robot did not have a camera or eyes; however, every single participant in our first study phase assumed the end-effector of the robot to be a head.}
\subsubsection{Results}

Each \textit{referent} had 16 resulting expressions. This results in $8 \times 16 = 128$ expressions. Which we reduced into the final expression set of 13 \textit{expressions}, see \autoref{tab:expressions}. Moreover, we classified each resulting expression based on our taxonomies into six categories. \autoref{fig:distributiontaxonomies} displays the distribution. 

\paragraph{Occurance Score ($OS$)}
We calculated the $OS$ for every expression, see \autoref{tab:os}. The mean $OS$ is 36.5 ($SD=15.42$). E10 received the highest $OS$ with 69\% while E08 received the lowest $OS$ with 12\%. 

\begin{table*}[t]
    \centering
    \caption{Calculated $OS$ (c.f., \autoref{sec:metrics}) in rounded percent (\%) for each \textit{expression-referent} combination. The same \textit{expression} could occur over different \textit{referents}. For example, from all the proposed expressions in R1 25\% were E01.}
    \label{tab:os}
    \small
    \begin{tabularx}{\linewidth}{Xlllllllllllll}
    \toprule
      & E01   & E02   & E03   & E04   & E05   & E06   & E07   & E08   & E09   & E10   & E11   & E12   & E13  \\
    \midrule
    \multirow{3}{*}{\shortstack[l]{$OS$}}   
                & R1=25 & R1=38 & R1=38 & R2=38 & R2=44  &R2=19 & R4=62 & R4=19 & R5=50 &  R7=69 & R7=31 & R8=56 & R8=44 \\
                & R3=38 & R3=25 & R3=38 &       &     &         & R5=38 & R5=12 & R6=38 &        &        &          &          \\
                & R4=19 &       &       &       &       &        & R6=50 & R6=12 &       &          &          &         &               \\
    \bottomrule
    \end{tabularx}
\end{table*}

\paragraph{Gesture Ratings}
We found that all six created curious expressions were perceived to be significantly more curious than the two control expressions and did not find any anomalies\footnote{See supplemental material for results}.  

\subsection{Example Expression Verification Study}
We continued the second study with our set of 13 \textit{expressions} (E01–E13). In the following, we describe the process of the expression verification phase. Please see videos of all expressions in the Supplementary Material.

\subsubsection{Study Design}
We conducted a between-subjects design online survey to confirm the gestures are understandable (N=260). Thus, each participant saw one video of the robot performing one expression (see \autoref{fig:expressionvideo}), interpreted the seen expression, and then completed a questionnaire. 

\subsubsection{Apparatus} 
\begin{figure}[t]
    \centering
    \begin{subfigure}{0.49\linewidth}
        \centering
    \includegraphics[width=\textwidth]{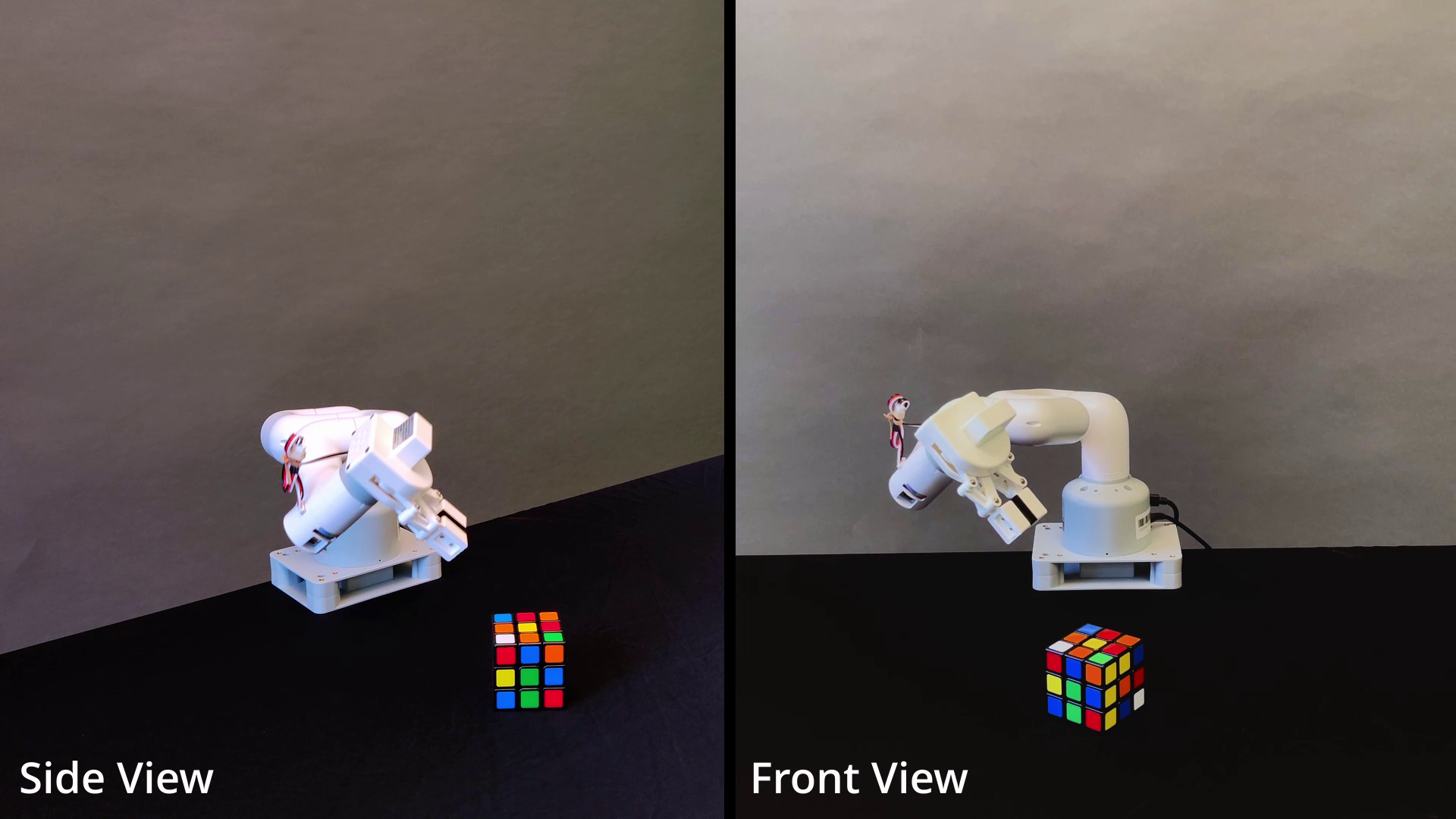}
        \caption{E02: Displaying the robot showing interest in an object.}
\label{fig:videos1}
    \end{subfigure}
    \hfill
    \begin{subfigure}{0.49\linewidth}
        \centering
        \includegraphics[width=\textwidth]{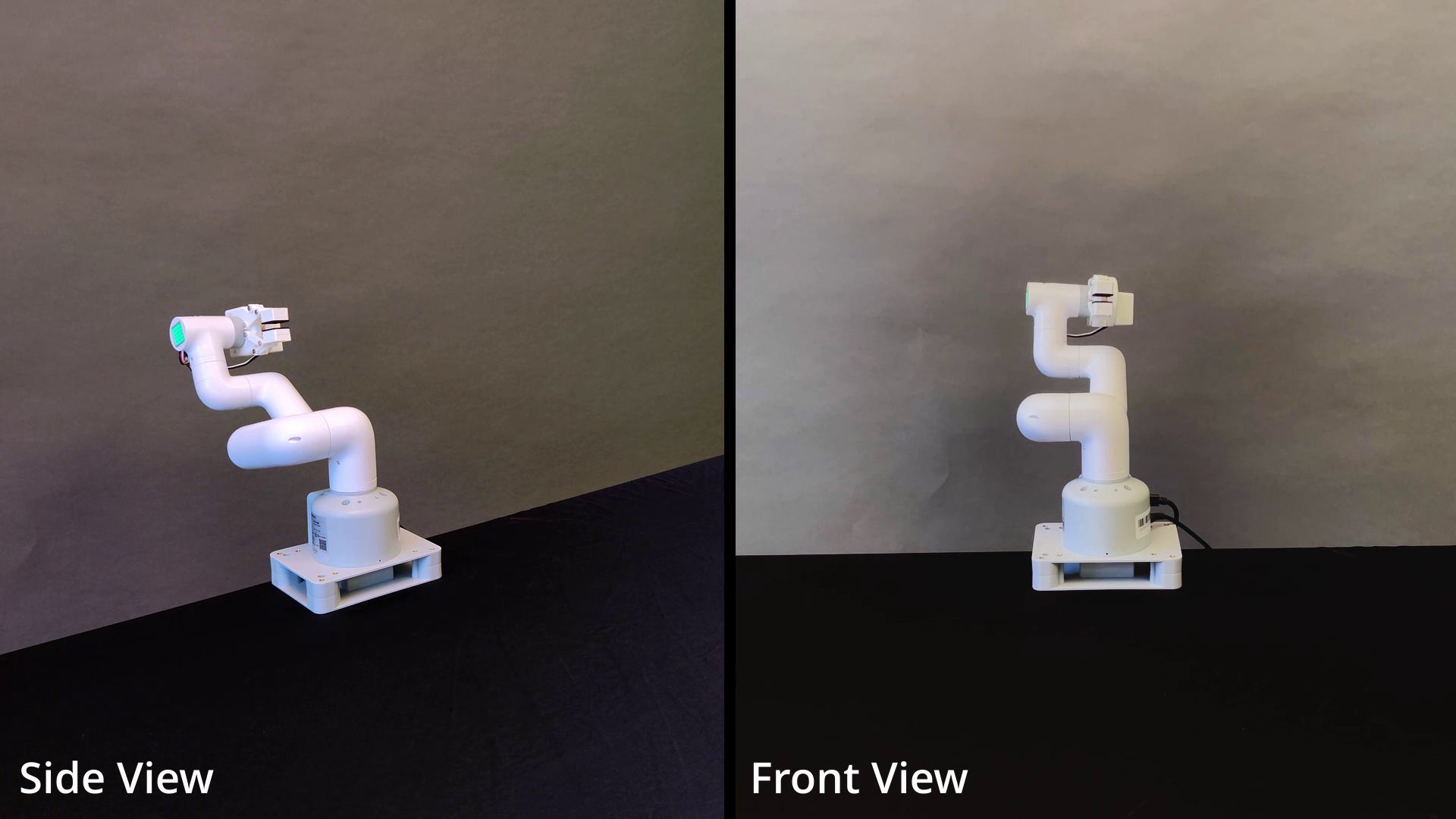}
        \caption{E10: Displaying the robot backing away in fear.}
    \label{fig:video2}
    \end{subfigure}
    \caption{Two example expressions from our setup to demonstrate the 13 expressions to the participants of the verification study in video form.}
    \Description{This figure shows two screenshots from the videos we showed to the participants of the second study. Both screenshots show the robot both from a side view (approximately 45 degrees) and a front view. The left screenshot depicts E02, where the robot is closely investigating a Rubix cube. The right screenshot shows E10. The robot is fully leaned back.}
    \label{fig:expressionvideo}
\end{figure}

To display the robotic expressions, we filmed each of the 13 expressions from two angles: the front, to simulate the participant's view, and the side, to better illustrate the depth of movement (see \autoref{fig:expressionvideo}). For expressions E01, E02, and E03, we included a Rubik's cube to show in which object the robot is interested. For the remaining expressions, the robot interacted directly towards the camera.

\subsubsection{Questionnaire}
We first asked participants to read an introduction to the robot's non-verbal communication and then watch a video of the robot performing an expression. We then asked them to describe what they believed the robot expressed briefly. We ensured participants had to watch the video by removing the button to continue the survey until the video ended. Next, participants rated the robot's movement using VAS sliders for ten questions, with sliders ranging from 0 (Strongly Disagree) to 100 (Strongly Agree) and no numbered ticks, as these have been shown to yield more precise responses~\cite{matejka2016effect,reips2008intervallevel}. Participants could rewatch the video as needed. We then asked to which extent the participant agreed with the following questions. (1) I perceived the gesture to be very engaged. (2) I perceived the gesture to be very attentive. (3) I perceived the gesture to be very explorative. (4) I perceived the gesture to be very information-seeking. (5) I perceived the gesture to be very curious. (6) This gesture is very understandable. (7) This gesture is very effective in communicating the robots' intent. (8) This gesture is very intrusive. (9) This gesture is very noticeable. (10) This gesture is very disturbing. 

\subsubsection{Participants}
We recruited 289 participants via Prolific\footnote{\url{https://www.prolific.co/}}, excluding 3 for failing attention checks and 26 for describing only the robot's movement instead of its interpretation\footnote{The participants were clearly instructed to state their interpretation and not describe movement}, leaving 260 participants (female = 131, male = 128, non-binary = 1). We ensured gender and nationality balance in our between-subject design, with each participant rating one of 13 expressions, resulting in 20 responses per expression. Participants, aged 19 to 68 ($M=34.60, SD=9.90$), included 209 full-time and 51 part-time employed individuals. They came from over 45 countries, with the highest numbers from the UK (18), Portugal (16), South Africa (15), Spain (14), Netherlands (13), Italy (12), Hungary (12), Ireland (11), Australia (10), Canada (10), Greece (10), and Mexico (10). On average, participants took 3.99 minutes ($SD=1.77$) to complete the survey and watched the video 2.98 times ($SD=3.33$).

\subsubsection{Analysis}
We utilized open coding for each expression description, iteratively refining the main component for each statement and reaching a consensus on a single label. However, some participants suggested that the robot expression could convey multiple meanings, leading to multiple labels within the final code for one expression. This process yielded 68 unique labels and 333 in total across 260 descriptions. Subsequently, we grouped these labels into 12 distinct code groups for further analysis. Detailed documentation of this coding process is available in the Supplemental Material.

\subsubsection{Results}
In the following, we report our results for the second study phase. This includes results regarding the expression descriptions and results of the questionnaires.

\begin{figure*}[t]
    \centering
    \includegraphics[width=\linewidth]{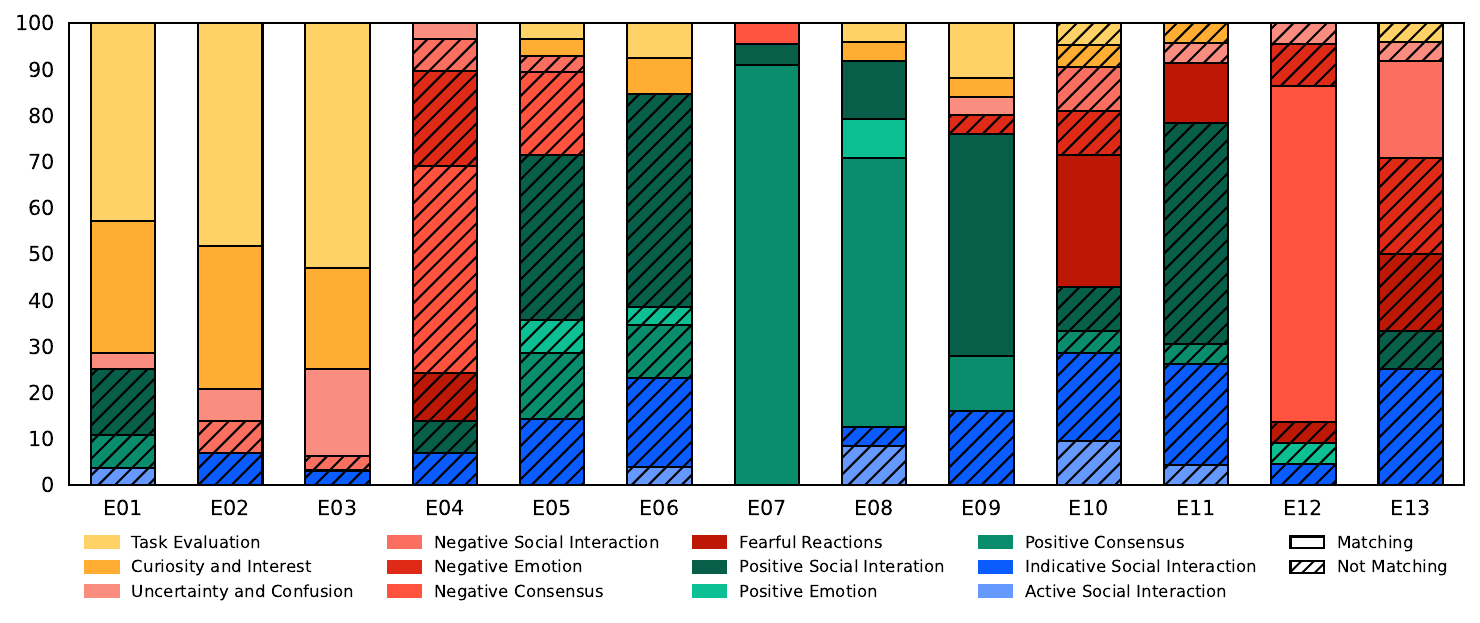}
    \caption{Distribution of groups of different interpretations for each expression. We striped non-matching expressions, c.f., \autoref{tab:expressionscores}.}
    \Description{This figure shows the distribution for each Expression (E01-E13) of labels. The four different themes we concluded from our coding process have four different colors, with gradients for the different subgroups. It shows, for example, that E01-E03 has mostly been described to be task evaluation or curiosity and interest. Furthermore, it shows whether this label fits the original idea of the referent. E01, for example, has a 75\% match. All these values are also in Table 3.}
    \label{fig:expressionlabels}
\end{figure*}

\begin{table*}[t]
\centering
\caption{Calculated qualitative response accuracy ($QRA$) (c.f., \autoref{sec:metrics}) in rounded percent (\%) for each \textit{expression-referent} combination. Each value is the percentage of how often an \textit{expression} was described with a matching label for a \textit{referent}. As multiple \textit{expressions} could arise from one \textit{referent}, one \textit{expression} has a separate matching score for each \textit{expression-referent} combination. 100\% means that every single label given by participants in the second study is a fitting label for its referent. One column can be read as follows: To express R1, the \textit{expression} E01 was correctly detected by 75\% of participants.}
\label{tab:expressionscores}
\small
\begin{tabularx}{\linewidth}{Xlllllllllllll}
\toprule
 & E01 & E02 & E03 & E04 & E05 & E06 & E07 & E08 & E09 & E10 & E11 & E12  & E13  \\
\midrule
\multirow{3}{*}{\shortstack[l]{$QRA$}}   
& R1=75 & R1=86 & R1=94 & R2=3 & R2=7 & R2=15 & R4=96 & R4=88 & R5=80 &  R7=29 & R7=13 & R8=73 & R8=21 \\
& R3=75 & R3=86 & R3=94 &         &         &          & R5=96 & R5=88 & R6=80 &          &         &          &          \\
&       &       &       &         &         &          & R6=100& R6=88 &          &          &         &          &          \\
\bottomrule
\end{tabularx}
\end{table*}

\paragraph{Expression Descriptions}
In total, we collected 333 labels from the 260 gesture descriptions across 68 unique labels. We found 12 overarching code groups, further categorized into four themes. \autoref{fig:expressionlabels} displays the distribution of the code groups for each expression. The main themes we found are (1) Exploratory Expressions, (2) Negative Expressions, (3) Positive Expressions, and (4) Interactive Expressions. The theme Exploratory Expressions include the labels \textit{task evaluation} ($52$) and \textit{curiosity and interest} ($31$). These groups include labels like ``assessing task'', ``examining'', or ``curiosity''. 
The Negative Expressions theme include the labels \textit{negative emotion}~($6$), \textit{negative social interaction}~($13$), \textit{negative consensus}~($8$), \textit{fearful reactions}~($17$), and \textit{uncertainty and confusion}~($14$). \textit{Negative emotion} includes labels like ``sadness,'' ``anger,'' or ``annoyment,'' \textit{negative social interaction} ``ignoring'' or ``distancing,'' and \textit{negative consensus} ``disagreement'' or ``unwillingness.'' The theme Positive Expressions includes the labels \textit{positive social interaction}~($59$), \textit{positive emotion}~($6$), and \textit{positive consensus}~($48$). \textit{Positive social interaction} expressions include labels like ``approaching,'' ``available,'' or ``submissive,'' positive emotions ``empathy'' or ``happiness'', and \textit{positive consensus} ``acceptance,'' ``agreement,'' or ``confirmation.'' Lastly, the theme Interactive Expressions includes the two labels \textit{active social interaction}~($7$) and \textit{indicative social interaction}~($35$). \textit{Active social interactions} include labels like ``dancing'' or `laughing'' and \textit{indicative social interactions} like ``commanding,'' ``taking,'' or ``pointing.''

We calculated $QRA$ between each expression and the goals of the \textit{referents}, displaying the results in \autoref{tab:expressionscores}. For instance, expression E01, which was created for referents R1 and R2, achieved a 75\% qualitative response accuracy, indicating that 75\% of labels matched the goals of the original referents. Expressions related to visual observation (R1, R3) and attentive listening (R4, R5, R6) were accurately identified by most participants. However, expressions for observing an unknown sound (R2) were correctly described by only 3\% and 7\% of participants, indicating poor suitability for expressing interest in sound. For the control referent expressing fear (R7), neither of the created expressions was correctly identified by most participants. Conversely, one expression (E12) for signaling disinterest performed well, while another (E13) did not.

\begin{figure*}[t!]
    \centering
    \includegraphics[width=\linewidth]{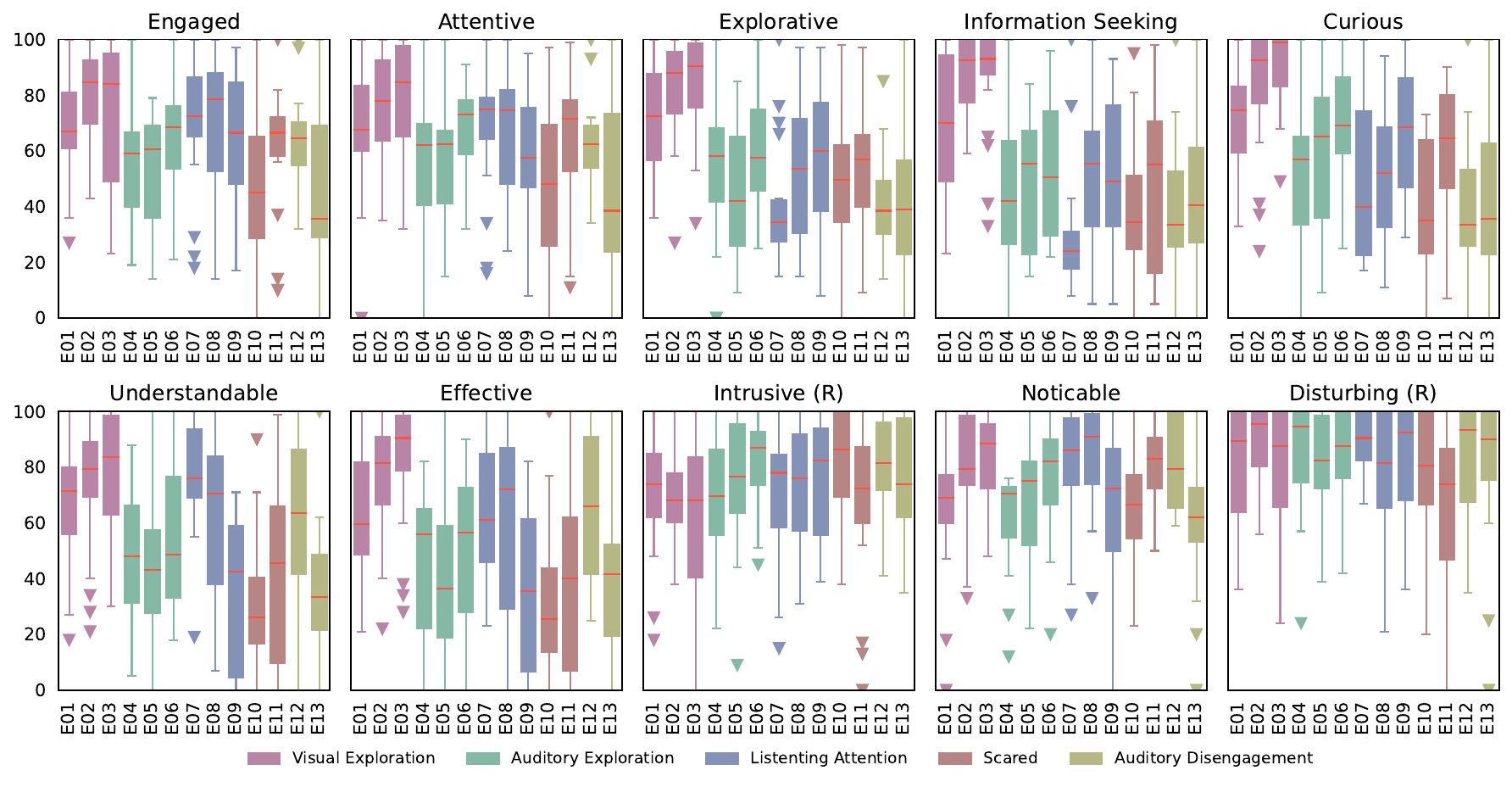}
    \caption{Boxplots of the ten questions we asked for every \textit{expression}. The top row displays the results for the curiosity questions. The bottom row displays the assessment of perception and impact questions. The scales for intrusive and disturbing are reversed such that 100 always means good.}
    \Description{This figure displays boxplots for the ten questions we asked per expression. The boxes are colored depending on one of 5 themes that emerged from our referents. It shows that E01-E03 has the highest values for engaged, attentive, explorative, information-seeking, curious, understandable, and effective. All expressions have high noticeable, nondisturbing, and intrusiveness scores.}
    \label{fig:quantresultsstudy2}
\end{figure*}

\begin{figure*}[t!]
    \centering
    \includegraphics[width=\linewidth]{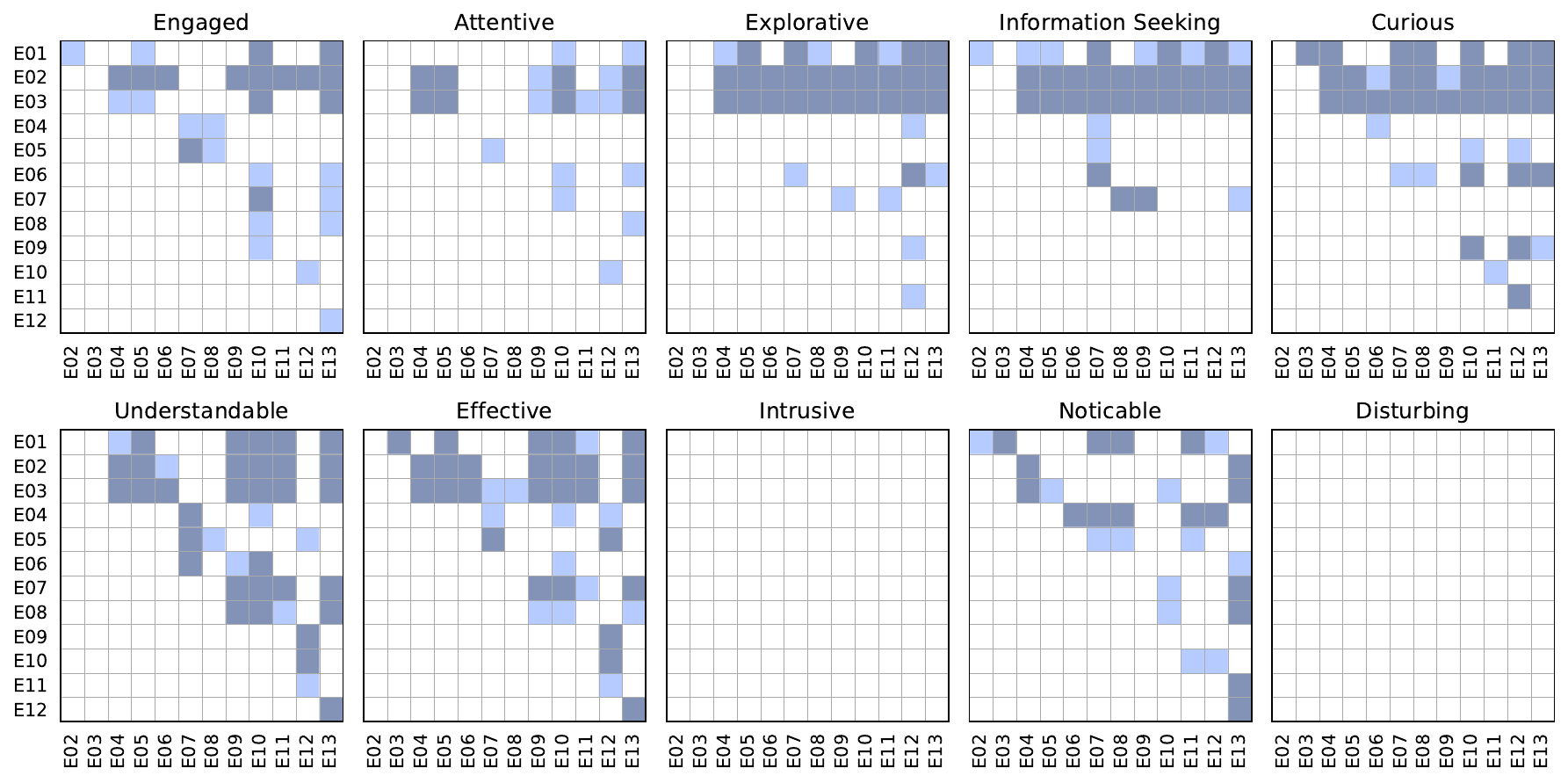}
    \caption{Significant differences calculated through a Mann-Whitney U test. Darker blue values show a significant difference with $p<0.01$, and lighter blue $p<0.05$. Empty grids show that this measurement had no significant effect. }
    \Description{This figure shows the results from our conducted post hoc tests on the 10 items from the questionnaire. Only two items (intrusive and disturbing) had no significant effect at all, thus, the grids are empty. For the other ones it shows many significant differences between the different expressions. The most important important ones we talk about in the text.}
    \label{fig:posthoc}
\end{figure*}

\paragraph{Curiosity and Perception Questionnaire}
The questionnaire results are shown in \autoref{fig:quantresultsstudy2}. Normality tests using the Shapiro-Wilk test revealed that none of the ten questions (five on user perception and impact, five on curiosity) followed a normal distribution ($p < 0.01$). Therefore, we used Kruskal-Wallis tests to detect significant effects among the expressions. We observed significant effects for all five curiosity-related questions and three out of five questions related to perception. Subsequently, we conducted Mann-Whitney U post hoc tests to determine the significant differences between expressions, which are presented in \autoref{fig:posthoc}.


\textbf{Visual Exploration:} For expressions E01, E02, and E03 (arising from R1 and R3), we found no significant difference for attentive, explorative, and understandable. We found E02 to be significantly more engaged and information-seeking than E01. We found E03 to be significantly more curious and effective than E01. And lastly, we found E02 and E03 to be significantly more noticeable than E01. 

\textbf{Auditory Exploration:} For expressions E04, E05, and E06 (arising from R2), we found no significant difference for engaged, attentive, explorative, information-seeking, understandable, and effective. We found E06 to be significantly more curious and noticeable than E04. 

\textbf{Listening Attention:} For expressions E07, E08, and E09 (arising from R4, R5, and R6), we found no significant difference for engaged, attentive, curious, or noticeable. We found E08 to be significantly more information-seeking than E07, E09 significantly more explorative and information-seeking than E07. However, we found E09 to be significantly less understandable and effective than E07 and E08.

\textbf{Scared:} For expressions E10 and E11 (arising from R7), we found no significant differences for engaged, attentive, information-seeking, understandable, and effective. We found E11 to be significantly more curious and noticeable than E10.

\textbf{Auditory Disengagement}: For expressions E12 and E13 (arising from R8), we found no significant differences for attentive, explorative, information-seeking, and curious. We found E13 to be significantly less engaged, understandable, effective, and noticeable than E12.

\section{Discussion}
In this paper, we proposed a new approach to create human understandable robot expressions via two phases. We then showed the applicability of our approach through an example study to create curious robotic expressions. In the following we will discuss our proposed approach.

\subsection{Towards a Robust Approach for Expression Elicitation and Validation}
Traditionally, developing expressive expressions for robots in HRI required laborious processes, including creating expressions from scratch, relying on domain experts, or using motion data of human users. However, these methods often lacked efficiency and were not suitable for all HRI scenarios. In our quest to create robot expressions that convey curiosity, we found that existing approaches did not align with our needs. Recognizing the growing importance of HRI research, we set out to establish a rapid and validated method for generating expressive robot expressions, addressing the need for efficiency and flexibility. In designing our approach, we established five key goals (see \autoref{sec:approachGoals}). 
\begin{enumerate}
    \item \textbf{Universal:} We demonstrated the adaptability of our approach for a non-humanoid robot arm. However, this can also be applied to other non-humanoid and also humanoid robots. A key strength of our example study was allowing participants to directly move the robot, and future studies should continue to prioritize ease of expression creation. Participants can manipulate the robot by hand, with tools, or by instructing a conductor, depending on the robot used. Regardless of the method, the expression generation process should be iterative, allowing users to refine expressions until they are satisfied.
    \item \textbf{Generalizable:} Our approach was designed to be generalizable by using flexible referents, allowing participants to create expressions for any topic. This topic-agnostic method is adaptable to different HRI contexts without relying on specific datasets or subject knowledge, making it applicable across various robots and scenarios. The use of \textit{referents} ensures consistency while allowing for diverse, transferable expressions that can be compared across studies.
    \item \textbf{Accessible:} Our method intentionally excluded the need for external datasets or domain experts, ensuring accessibility to all researchers. However, this does not mean that experts can not be used for the expression elicitation phase. 
    \item \textbf{Comparable:} By employing taxonomies and qualitative response accuracies, we aimed to provide a clear description of the created expressions and their comprehensibility, making it possible to compare results across studies and generalize findings.
    \item \textbf{Validated:} To confirm the understandability of the expressions, we propose conducting an online study with a large participant pool. This approach ensures broad feedback, helping confirm that the created expressions are comprehensible to a diverse audience.
\end{enumerate}
In summary, our approach successfully met all five goals, offering a versatile and efficient method for generating robot expressions in HRI research.

\subsection{Our Approach Successfully Uses Human Creativity to Design Understandable Expressions}
For the expression creation part, we used ideas from gesture elicitation studies and combined them with enactment. We found that this second part really got participants in our example study into the correct thinking mood in which they would express the different \textit{referents} and, thus, how they would want the robot to express these things. This worked very well in our case, even though we chose a non-humanoid robot, using the difficult part of the spectrum of user-created expressions as there is no direct 1-to-1 mapping between their own expressions and how the robot should express that, and some abstraction is needed. We propose that as this approach worked for a non-humanoid robot, using this approach for a robot with more humanoid or animalistic traits will surely work, as participants can use even more previous knowledge to create expressions. 

We aimed to create expressions in the domain of curiosity and successfully created six expressions. With the knowledge of which core ideas from expressions lead to participants assuming the robot expresses something designers can now take these ideas and design choreographies for robots. In our expression validation phase, we did not give the users any context about the robot's current routine. However, as human expressions are interpreted differently, depending on the context~\cite{mcneill2005gesture, poggi2010types}, it is important to note that also different interpretations of the same expression should be labeled as correctly understood. By giving users more context, the potential interpretation of the robot's expression can also be limited, which might be desirable or undesirable. 

\subsection{Generalizablility Comes Across Studies Not Within}
In our example study, we intentionally designed the study parameters to maximize generalizability and avoid priming participants. We refrained from specifying the direction in which robot expressions should be performed, allowing participants to choose their reference point for the robot's ``front.'' Notably, we observed that expressions with direct contextual cues performed better, as seen in the comparison between visual and auditory exploration, c.f., \autoref{fig:quantresultsstudy2}.

Despite these challenges, we maintained our commitment to generalizability. We suggest that future studies employing this method consider more specific guidelines to streamline the expression creation process. Providing clearer information does not necessarily limit creativity but can facilitate researchers' understanding of participants' core ideas while preserving expression variability. This approach would enhance the method's efficiency and reproducibility across multiple studies. Many now-established HCI methods have gone through this iterative process of getting improved and also verified over time through the HCI community, e.g., general gesture elicitation studies~\cite{wobbrock2009Userdefined}, the NASA-TLX questionnaire~\cite{hart1988development, hart2006nasatask}, or ART Anovas~\cite{wobbrock2011aligned}. We aimed to create a generally applicable approach to create human-created robotic expressions and validate their understandability in a rapid way, as many HRI studies need robotic behavior. We also note that our approach only generates verified general expressions that are understandable. In contrast to other works, we do not focus on the design process of these gestures, which focus on generating ``crisp'' looking expressions~\cite{koike2023exploringa}. We propose that this should be done by designers when applied afterward.

\subsection{Cultural-dependence of Robot Expressions}
Non-verbal communication varies significantly across different cultures. For example, while Western countries commonly use nodding expressions to signify agreement, in Bulgaria, this same expression is interpreted as disagreement~\cite{andonova2012nodding}. Our approach to creating robot expressions directly reflects this cultural sensitivity by providing a one-to-one mapping of expressions based on the cultural backgrounds of the participants involved.

Human-human interaction encompasses a diverse array of cultural differences, and these distinctions are not universally transferable between cultures. However, when we tailor our approach by selecting participants from specific cultural backgrounds, the resulting set of expressions becomes directly applicable and understandable within that culture. Furthermore, by conducting similar studies with participant groups from different cultures, we can identify expressions that overlap and transcend cultural boundaries.

Considering this approach, we can design non-verbal robot communication systems that are culturally unbiased and adaptable to various cultural contexts, ensuring that the robot's expressions are universally comprehensible and respectful of cultural nuances.

\subsection{How and Who Does Our Approach Help}
Non-verbal communication holds immense importance in both human-human interactions and human-robot interactions. Implicitly conveying one's current state and needs through non-verbal cues is particularly valuable in human-robot communication because it seamlessly integrates with ongoing tasks, unlike intrusive notifications from smart devices, which can disrupt the user's focus. With robots, we have the unique opportunity to redefine how we interact with advanced technical systems capable of providing comprehensive support in our daily lives. Achieving seamless and effective communication is pivotal for robots to fulfill their potential as supportive companions.

However, traditional methods of developing non-verbal communication for robots have limitations and are mostly data-driven. Here, we want to add the the options of how future robot expressions can be created through a more human-centered approach. Our approach is usable with both expert and non-expert users for the expression elicitation phase, depending on the domain for which the expressions should be created for. Through the validation phase with a large-scale sample of non-experts, it becomes clear which expressions will be understood by the general population. 

\subsection{Insights from our Example Study}
In the following, we will discuss our findings from our example study to create human-understandable robotic expressions expressing curiosity. 

\subsubsection{People Elicit Convergent Expressions}
In our expression elicitation phase, we observed that participants often shared similar ideas for expressing various \textit{referents}. Out of 128 recorded expressions, we identified 13 unique expressions, averaging $9.84$ expressions per type ($SD=5.35$). This similarity highlights that a large participant pool may not be necessary due to the low variation in human expressions. Providing similar referents allowed participants to be creative while addressing comparable ideas. Human expressions tend to exhibit similarities across various situations, with contextual cues playing a vital role in interpreting the intended meaning of an expression. This contextual influence is further evidenced by the consistently higher scores for expressions E01, E02, and E03 across all measurements in the second phase, as these expressions were the only ones framed within the context of an object.

\subsubsection{Human-Understandable Curious Expressions for Visual Exploration and Listening Attention}
Our study revealed that of the 13 human-created expressions, seven were well understood by participants, while six were not. This indicates that we successfully created at least one well-understood and, thus, effective expression for six of the eight referents (five of six curiosity referents). We can now compare these effective expressions with the quantitative questionnaire results.

For referents R1 and R3, expressions E01, E02, and E03, which had high qualitative response accuracies, were all suitable for conveying visual interest. However, E03 had the highest accuracy of 94\%, making it the best match for R1 and R3. It combines both continuous and discrete flow, distinguishing it from E01 and E02, which use only one flow type. For referents R4, R5, and R6, expressions E07, E08, and E09 were applicable. E07 had the highest accuracy (96\% for R4 and R5, 100\% for R6), while E08 and E09 were deemed significantly more information-seeking than E07, though E09 was less understandable. E08, with an 88\% accuracy, and E07 are both suitable for these referents. Given the similarity in their movements (nodding with varying body involvement), these results align with expectations. Referent R8 had two matching expressions (E12 and E13). E12 had a higher qualitative response accuracy (73\%) compared to E13 (21\%) and was significantly more engaged, understandable, effective, and noticeable. Consequently, E12 emerged as the superior choice for expressing robot rejection of a person's statement. Thus, head shaking is more understandable than turning the body away.

This shows that we successfully created human-understandable expressions for visual curiosity (E1, E2, E3) and conversational curiosity (E07, E08, E09). These expressions effectively conveyed the robot's interest in visual objects and attentiveness to spoken conversation. However, auditory curiosity expressions faced challenges and were poorly understood by participants. The effective visual curiosity expressions involved the robot approaching the object, inspecting it from various angles, and using its entire body for movement. In contrast, effective listening attention expressions primarily featured nodding with the robot's end-effector, while another expression involved body movement, shifting from facing away to facing and approaching the users. We attribute the success of these expressions to (1) the contextual relevance of the target object and (2) the clarity of expressions associated with listening attention, such as nodding. Auditory exploration expressions lacked these advantageous properties, contributing to their lower participant comprehension rates.

\subsubsection{Obvious Expressions are Understandable and Context Matters}
Our study demonstrated that nodding and head shaking effectively signal interactions with the robot, whether inviting or rejecting conversations. These findings align with human behavior, where a limited set of expressions is employed to convey various emotional states~\cite{barrett2019emotional}, emphasizing the importance of familiar and comprehensible expressions. Additionally, we found that even with a non-humanoid robot, participants generated human-like expressions, and visual curiosity gestures—such as examining objects—were well understood. Expressions indicating attentiveness to spoken conversation also led to high participant comprehension.

However, the study revealed challenges in participants' understanding of auditory curiosity expressions. We attributed this confusion to the absence of context in the gestures. By filming the robot's expressions without providing contextual information, participants struggled to interpret the intended meaning behind auditory curiosity expressions. This highlights the significance of incorporating context into expressions, which can substantially enhance their interpretability in human-robot interactions.

\subsection{Limitations and Legacy Bias}
The issue of ``legacy bias'' in gesture elicitation studies, as highlighted by \citet{morris2014reducing}, raises a valid point. It suggests that when participants are asked to create gestures for novel interaction capabilities, well-known and familiar gestures tend to emerge as the preferred choices. For instance, when prompted to create a gesture for zooming, many participants might instinctively choose a pinching motion, reflecting their familiarity with touchscreens.

While this legacy bias might initially seem limiting, it does not necessarily have to be viewed as a negative aspect. Human-computer interaction has evolved over decades, and users have become accustomed to certain interaction concepts. Requiring users to relearn entirely new ways of interacting with technical devices each time a novel interaction capability emerges could be cumbersome and counterproductive.

In this context, our method, which prioritizes well-known and obvious expressions, can be seen as a valuable tool. It allows researchers to systematically identify and leverage the most common and widely understood non-verbal communication cues. These cues are already ingrained in users' minds due to their experience with various technical devices. Thus, our approach enables researchers to tap into this existing knowledge base and build upon well-established interaction concepts, making the transition to new technology more seamless and intuitive for users.

\section{Conclusion}
In this work, we presented a new approach to elicit and generate human-understandable robot expressions. This helps to create better human-robot interaction as a better understanding of robots' intentions leads to more seamless collaboration. Our approach uses ideas from gesture elicitation studies, human creativity and intuition, and wide verification. Our approach is generalizable; using it, creating expressions for all variations of different robots is possible. We showed our approach with an example case of creating curious robot expressions with a non-humanoid robot. We have successfully created expressions to deliver five out of six referents correctly. We found that robot expressions, in contrast to interaction gestures, do not have to be distinctly different from each other, and the same expressions can be the best way to convey different robot states. Through our study, we found that familiar expressions tend to be understood better than novel expressions. With this approach, we aim to support future work for HRI; through a way to easily and rapidly fabricate understandable robotic expressions. 

\bibliographystyle{ACM-Reference-Format}
\bibliography{main_Arxiv}

\end{document}